\pdfoutput=1
\documentclass[11pt,a4paper]{article}
\usepackage[hyperref]{acl2021}
\usepackage{times}
\usepackage{latexsym}

\usepackage{booktabs}
\usepackage{amssymb}
\usepackage{pifont}
\newcommand{\cmark}{\ding{51}}%
\newcommand{\xmark}{\ding{55}}%
\newcommand{\fdia}{\lozenge}
\newcommand{\gbox}{\Box}

\usepackage{microtype}

\aclfinalcopy 

\title{Structural Ambiguity and its Disambiguation in Language Model Based Parsers: the Case of Dutch Clause Relativization}

\author{Gijs Wijnholds \\
  Institute of Language Sciences \\
  Utrecht University \\
  \texttt{g.j.wijnholds@uu.nl} \\\And
  Michael Moortgat \\
  Institute of Language Sciences \\
  Utrecht University \\
  \texttt{m.moortgat@uu.nl}\\}

\date{}

\begin{document}
\maketitle
\begin{abstract}
This paper addresses structural ambiguity in Dutch relative clauses. By investigating the task of \emph{disambiguation by grounding}, we study how the presence of a prior sentence can resolve relative clause ambiguities. We apply this method to two parsing architectures in an attempt to demystify the parsing and language model components of two present-day neural parsers. Results show that a neurosymbolic parser, based on proof nets, is more open to data bias correction than an approach based on universal dependencies, although both set\-ups suffer from a comparable initial data bias.
\end{abstract}


\section{Introduction}
Ambiguity pervades natural language and as such forms one of the central challenges for natural language understanding (NLU) systems. Given the fact that most such systems rely on large-scale deep learning architectures, the presence of structural biases in the training data used may affect a system's capacity for disambiguation. Specifically in the case of parsing, typical architectures rely on the assumption of just a single correct parse, although many may exist. This assumption then may force a bias into the training, both on the lexical and on the syntactic level.

In this paper, we study syntactic ambiguities in Dutch, where a structural ambiguity affects the interpretation of relative clauses. The preferred reading as subject or object relativisation will typically be determined by lexical choice. A running example in Dutch, with its two possible interpretations in English, is given below:
\begin{center}
    \begin{tabular}{cl}
        $(a)$ & de dokter die de pati\"{e}nt geneest \\
        $(b)$ & \emph{the doctor who cured the patient} \\
        $(c)$ & \emph{the doctor whom the patient cured} \\
    \end{tabular}
\end{center}

In this example, the verb `cure' displays a strong selectional preference for a doctor as its subject and a patient as its object, so one would expect that the subject-relative interpretation $(b)$ is the preferred reading. However, grounding the phrase in a  prior sentence disambiguates the ambiguous relative clause: by prepending the phrase ``The patient cured the doctor", one is led to infer that the object-relative reading of $(c)$ is in fact preferred. Orthogonally, there are cases of lexical choice that \emph{block} one of the readings, typically due to the semantic class of the subject and object being different. An example is ``De man die de boterham eet" (\emph{The man who eats the sandwich}), where it is semantically implausible for the verb's arguments to be reversed. The situation is summarized in Table \ref{table:examples}, where different possible orderings of the relative clause together with a prior sentence lead to a different expected readings of the relative clause.


Arguing that a parser typically exploits statistical properties of its training corpus, but should additionally rely on both lexical and syntactic cues inside of that corpus, we carry out an experiment test a parser's capacity for disambiguation in context. Specifically, we extract a set of selectional preferences for Dutch transitive verbs, that we classify according to their \emph{reversibility}, i.e. whether subject and object can be interchanged, additionally indicating whether there is a strong preference for a noun as subject or object. This leads to three classes of $(s, v, o)$ triples that we then use to generate a test set of Dutch relative clauses together with prior sentences, to test a parser's capability of disambiguation in context.

By evaluating two different parsing regimes that both are built on top of a language model, we investigate the encoding of structural bias in the parser training data, the possibility of mitigating structural bias, and attempt to pinpoint to what extent lexical knowledge or syntactic information is employed by the parsers in question.

\begin{table*}[h!]
    \centering
    \begin{tabular}{@{}llll@{}}
        \toprule
         \textbf{Prior sentence} & \textbf{Target phrase} & \textbf{Correct reading} & \textbf{Plausible} \\
         \midrule
         De man eet de boterham. & De man die de boterham eet & subj. rel. & \cmark \\
         De man eet de boterham. & De boterham die de man eet & obj. rel. & \cmark \\
         De boterham eet de man. & De man die de boterham eet & n/a & \xmark \\
         \midrule
         De dokter geneest de patient. & De dokter die de patient geneest & subj. rel. & \cmark \\
         De patient geneest de dokter. & De dokter die de patient geneest & obj. rel. & \cmark \\
         De dokter geneest de patient. & De patient die de dokter geneest & obj. rel. & \cmark \\
         De patient geneest de dokter. & De patient die de dokter geneest & subj. rel. & \cmark \\
         \bottomrule
    \end{tabular}
    \caption{Different cases in our disambiguation experiment, where the provided prior sentence determines the interpretation of the target phrase. The top three rows are cases of an \emph{irreversible} $(s, v, o)$ triple where interchanging subject and object leads to an implausible case that is not included in our experiments. The bottom four rows are cases of a \emph{reversible} $(s, v, o)$, albeit with a strong selectional preference for one interpretation. By adding the prior sentence, the interpretation of the target phrase is disambiguated.}
    \label{table:examples}
\end{table*}

Our contributions in this paper are therefore threefold: we (1) provide a dataset of selectional preferences for Dutch $(s, v, o)$ triples with an additional layer of classification according to semantic noun classes, and (2) create a novel test set targeting structural ambiguity in the interpretation of Dutch relative clauses. Finally, we (3) provide a number of experiments indicating that structural bias is easily encoded, but not so easily mitigated in a language model-based parser.

\section{Background}

\paragraph{Probing and syntactic sensitivity} Previous work has used probing, where a small neural network is attached to a large language model to extract task-specific information, to argue that large scale language models like BERT have internalized some linguistic knowledge during pretraining \cite{tenney2018what}, and there appears to be some consensus of the syntactic awareness of BERT models \cite{rogers-etal-2020-primer}. Specifically, studies have indicated the possibility of extracting parse trees from BERT representations succesfully \cite{hewitt-manning-2019-structural,vilares2020parsing}.

Another line of research into syntactic sensitivity investigates the probabilities of language models in a masked language modelling environment, where studies typically define surprisal rates to measure the degree to which the language model's predictions coincide with human-like behaviour in the face of syntactic ambiguities \cite{futrell-etal-2019-neural,hu-etal-2020-systematic,arehalli-etal-2022-syntactic,aina-linzen-2021-language}, typically focusing on garden-path effects. A related approach uses priming to investigate the language models' response to structurally similar sentences \cite{10.1162/tacl_a_00504}.

These studies differ from the current paper in that we explicitly target sentences that are not disambiguated without adding extrasentential context, which should for a parser to infer the intended syntactic analysis. Hence, our setup relies on a probing-like paradigm where we evaluate a parser on top of a language model.

\paragraph{Dutch NLP} The rising interest in large scale language models in the NLP community has led to a number of investigations for Dutch specifically. Two dominant Dutch language models have been developed, based on the respective BERT~\cite{devlin-etal-2019-bert} (BERTje, \citet{de2019bertje}) and the RoBERTa~\cite{liu2019roberta} architecture (RobBERT, \citet{delobelle-etal-2020-robbert}). On the evaluation side there have been several studies using Dutch-specific phenomena to evaluate the respective monolingual language models: the work of \cite{wijnholds-moortgat-2021-sick} introduces a parallel Natural Language Inference (NLI) dataset for Dutch, showing that Dutch NLI is more difficult to tackle than its original English version. More recently, a range of probing studies has been performed investigating verb-subject dependencies in the face of syntactic constructions involving discontinuity \cite{kogkalidis-wijnholds-2022-discontinuous} and ellipsis \cite{Haagen_Dona_Bosscha_Zamith_Koetschruyter_Wijnholds_2022}, aside from a more subtle Dutch NLI challenge \cite{wijnholds-2023-assessing}.

More closely related to constructions involving relative pronouns is the work of \citet{allein2020binary}, which introduces relative pronoun prediction for Dutch \emph{die} and \emph{dat} as a binary classification task, where the surrounding context determines the choice of the neuter (\emph{dat}) or non-neuter (\emph{die}) pronoun. This can be modelled as a masked language modelling task \cite{delobelle-etal-2020-robbert}, reaching high performance. A more complex variation of this task is defined by \citet{Bouma_2021}, where the experiment investigates the language model's capacity to predict relative pronoun attachment.

Against these studies, our approach differs in that we do not directly target the language model probabilities, but rather investigate the attached parsers, thereby indirectly assessing the contextualization power of the underlying language model.
In that sense, the current work is more in line with prior, more theoretical work that takes parser ambiguity into account \cite{moortgat2017lexical,moortgat2020frobenius}. 

Finally, while previous work tried to exploit selectional preferences to improve parser accuracy \cite{van2010self}, we rather use selectional preference to investigate the bias of existing parsers, as a precursor to potential architectural considerations, in a setting where language models are commonly employed in parser development. We make the code for the different components available in three separate repositories.\footnote{\url{https://github.com/gijswijnholds/syntactic_nl2i}} \footnote{\url{https://github.com/gijswijnholds/dynamic-proof-nets-disambiguation}}\footnote{\url{https://github.com/gijswijnholds/udparsing_bert}}

\section{Data Generation}

In order to generate a suitably large set of relative clause patterns, we proceed with a pipeline for extracting suitable subject-verb-object triples from a large corpus. First we perform a probability-based extraction of base triples. Then we use lexical information from a dictionary to filter out relevant nouns and classify them according to their semantic class, after which in a final step we perform a manual filtering and classification of the obtained subject-verb-object triples to guarantee correctness.

\paragraph*{Triple extraction}

For the first step, we need a way to extract subject-verb-object triples from a large corpus. To this end, we iterate over Lassy Large \cite{vanNoord2013} -- a 700M word corpus with automatically assigned syntactic annotations -- and extract all the cases of transitive verbs and their respective subjects and objects.

To compensate for parsing errors and infrequent observations, we rely on posterior probability \cite{resnik-1997-selectional}, the simplest measure that was shown to give best performance on a selectional preference acquisition task \cite{zhang-etal-2019-sp}. Posterior probability allows us to easily filter out those triples that occurred most frequently, allowing us to consider the most canonical triples only. For a given $(s, v, o)$ triple, its posterior probability is defined as follows:
    \[p(s, v, o) = \frac{f(s, v, o)}{\sum\limits_{s',o'} f(s', v, o')}\]

\paragraph*{Dictionary-based classification}

After gathering all $(s, v, o)$ triples and removing stopwords, we perform a dictionary-based filtering and classification in two steps. First, we obtain an exhaustive list of Dutch nouns with their respective semantic class\footnote{\emph{person, animal, plant, substance, object, abstract, mass noun}} from the \emph{Algemeen Nederlands Woordenboek}\footnote{\url{https://anw.ivdnt.org}}, a comprehensive online dictionary of Dutch.

We use these categories to classify all the subjects and objects in the extracted $(s, v, o)$ triples, after which we organize triples according to the most frequent noun categories, preventing as much as possible infrequent or illicit combinations of a verb with a given pair of nouns. By organizing triples by whether subject and object fall into the same semantic category, we have an initial estimate of whether a triple is \emph{irreversible}, and in cases where the semantic category of the subject and object coincide, we estimate how strong the preference of the verb for the particular subject and object is based on the frequency of the $(s, v, o)$ triple relative to its inverted $(o, v, s)$ triple.

\paragraph*{Manual filtering} This initial estimation gives us a comprehensive list of subject-verb-object triples, that we finally filter manually, making a per case decision on whether a triple is \emph{irreversible}, reversible with \emph{strong} preference for the given subject and object, or reversible with a \emph{weak} such preference, meaning that object and subject could be swapped without leading to an implausible triple. An example of the latter is ``De toerist herkent de reiziger" (\emph{The tourist recognizes the traveller}).

\paragraph{Generating relative clauses}

After the two-step process described above, we have a robust set of $(s, v, o)$ triples that we can use to generate the desired test cases for our experiment. In total, we obtained 3304 irreversible triples, 370 triples with a strong preference for the regular relation, and 724 where subject and object could be easily interchanged.

From these triples, we generate relative clauses following the pattern displayed in Table \ref{table:examples}. First, for an irreversible $(s, v, o)$ triple, we create a relative clause with the subject $s$ as the head noun (\textbf{S Pron O V}) as well as the reversed relative clause (\textbf{O Pron S V}), which due to the irreversibility of the $(s, v, o)$ triple must be analyzed with the subject-relative and object-relative reading respectively. We prepend in both cases a prior sentence in SVO order to be able to inspect the effect of adding this context to the parser's input.

For the reversible triples, we generate four cases. Again, we generate two variations of the relative clause, but additionally vary the prior sentence so it will force a reading of the relative clause, that is ambiguous without this context. This allows us to compare parser performance on both lexical and syntactic cues.

In the generation process, we use the \emph{Algemeen Nederlands Woordenboek} to extract the gender of each noun (\emph{de} for gendered nouns, \emph{het} for neuter nouns), and the gender of the relative pronoun (\emph{die} for gendered head nouns, \emph{dat} for neuter head nouns).  

\section{Parsing Regimes}

In our experiments we want to evaluate parsers that were built on top of a large language model, in order to distinguish the effect of the language model from that of the specific parsing strategy employed. For comparison purposes we test two different parsing regimes.

\paragraph{Neural proof nets} The first parser we examine is a neurosymbolic parser based on a multi-modal type-logical grammar that simultaneously encodes function-argument structure and dependency roles \cite{spindle,kogkalidis-etal-2020-neural}. This parsing setup exists alongside other \emph{neuralizations} of categorial grammar parsers \cite{clark2021oldnew}, but was explicitly developed for Dutch.

The architecture of this parser follows the typical structure of a categorial parser, where a \emph{supertagging} component assigns to words logical formulas that encode their intended combinatorial behaviour, followed by a process of \emph{proof search} that combines the formulas into a proof representing the full parse history. In the system of \cite{spindle}, supertagging is implemented as a graph decoding network, that learns to construct the tree structures representing the formulas of the logical formalism from an underlying (Dutch) BERT model \cite{de2019bertje}. Proof search is implemented as \emph{neural proof net} search, 
which amounts to linking atomic subformulas of opposite polarity in a way that determines the correct dependency and function-argument relations.

For a full exposition of this parser we refer the reader to \cite{kogkalidis-etal-2020-neural}; for the sake of our experiments it is enough to consider the two possible correct supertagging assignments for the Dutch relative clause, illustrated in Table \ref{table:correct_parses_npn}.

\paragraph{Universal dependencies} The second parser we evaluate approaches parsing as a sequence labelling task, following the work of \citet{strzyz-etal-2019-viable}. 

\begin{table*}[h!]
    \centering
    \begin{tabular}{@{}lc@{\hskip 0.7em}ccc@{\hskip 0.7em}c@{\hskip 0.7em}c@{}}
                & {\footnotesize De} & {\footnotesize pati\"{e}nt} & {\footnotesize die} & {\footnotesize de} & {\footnotesize dokter} & {\footnotesize geneest} \\
        {\footnotesize Subj. rel.} & {\tiny $\gbox_{det}(\textsf{N} \multimap \textsf{NP})$} & {\tiny $\textsf{N}$} & {\tiny $\fdia_{relcl}(\fdia_{su}\textsf{VNW} \multimap \textsf{S}) \multimap \gbox_{mod}(\textsf{NP} \multimap \textsf{NP})$}& {\tiny $\gbox_{det}(\textsf{N} \multimap \textsf{NP})$} & {\tiny $\textsf{N}$} & {\tiny $\fdia_{obj1} \textsf{NP} \multimap \fdia_{su} \textsf{VNW} \multimap \textsf{S}$} \\
        {\footnotesize Obj. rel.} & {\tiny $\gbox_{det}(\textsf{N} \multimap \textsf{NP})$} & {\tiny $\textsf{N}$} & {\tiny $\fdia_{relcl}(\fdia_{obj1}\textsf{VNW} \multimap \textsf{S}) \multimap \gbox_{mod}(\textsf{NP} \multimap \textsf{NP})$}& {\tiny $\gbox_{det}(\textsf{N} \multimap \textsf{NP})$} & {\tiny $\textsf{N}$} & {\tiny $\fdia_{obj1} \textsf{NP} \multimap \fdia_{su} \textsf{VNW} \multimap \textsf{S}$} \\
    \end{tabular}
      \caption{Supertagging assignment of the neural proof net parser for the subject-relative and object-relative interpretation of the Dutch relative clause. Function-argument structure is encoded by the linear implication, where $A\multimap B$ denotes a function consuming a phrase of type $A$ to produce a result of type $B$. The unary operations $\fdia_d,\gbox_d$ encode dependency structure, where heads assign dependency role $d$ to their \emph{complements} by means of $\fdia_d$ marking, and $\gbox_d$ allows \emph{adjuncts} to project their dependency role $d$.
      The parser assigns a \emph{higher-order} formula to the relative pronoun, allowing the implicit gap for either the subject or object of the verb (``geneest") in the body of the relative clause to be identified with the head noun (``de pati\"{e}nt"). For the sake or our experiments, we can inspect the formula assigned to the relative pronoun `die' to determine the interpretation of the relative clause, given the annotation of the gap type \textsf{VNW} with either the $su$ or $obj1$ dependency.}
    \label{table:correct_parses_npn}
\end{table*}

\begin{table*}[h!]
    \centering
    \begin{tabular}{lcccccc}
                & {\footnotesize De} & {\footnotesize pati\"{e}nt} & {\footnotesize die} & {\footnotesize de} & {\footnotesize dokter} & {\footnotesize geneest} \\
                \midrule
    Subj. rel.  & {\footnotesize (\textsc{+1,N},$det$)} & {\footnotesize (\textsc{-1,ROOT}, $root$)} & {\footnotesize \textsc{(+1,V},$nsubj$)} & {\footnotesize (\textsc{+1,N},$det$)} & {\footnotesize (\textsc{+1,V},$obj$)} & {\footnotesize (\textsc{-2,N},$acl$:$relcl$)} \\
    Obj. rel    & {\footnotesize (\textsc{+1,N},$det$)} & {\footnotesize (\textsc{-1,ROOT}, $root$)} & {\footnotesize \textsc{(+1,V},$obj$)} & {\footnotesize (\textsc{+1,N},$det$)} & {\footnotesize (\textsc{+1,V},$nsubj$)} & {\footnotesize (\textsc{-2,N},$acl$:$relcl$)} \\
    \end{tabular}
    \caption{Relative part-of-speech based encoding of the subject-relative and object-relative interpretation of the Dutch relative clause. In both cases, the label assigned to the relative pronoun `die' determines the interpretation of the relative clause.}
    \label{table:correct_parses_ud}
\end{table*}

Specifically, sentences are encoded using a relative part-of-speech based encoding, with each word assigned a triple $(i, p, d)$ where $p$ refers to the part-of-speech of the word's head, $i$ indicating its relative location, and $d$ the word's dependency label. For example, the label (\textsf{+1, V}, $nsubj$) says that the current word is in the \emph{nsubj} dependency relation with respect to the first word to its right that carries the \textsf{V} part-of-speech tag. The root is encoded by labelling its dependent with (\textsf{-1, ROOT}, $root$). A full example is given in Table \ref{table:correct_parses_ud} which contains the encoding of the two possible parses for the Dutch relative clause. The relative part-of-speech based encoding was found to be the highest performing in the experiments of \citet{strzyz-etal-2019-viable} and so we use it in our experiments. Labels are considered atomically, and as such the parser is implemented as a standard token classification model, where a token-level classifier is fine-tuned along with a BERT model.

\section{Evaluation \& Results}

In our experiments, we test the parsers on three different scenarios: first, we inspect the existing structural bias present in the parsers due to training data statistics, in a setting where the parser only gets fed the relative clause. In such a setting, we would expect the parser to have a strong bias for irreversible cases, but an even distribution in accuracy on weakly reversible cases. Next, we observe the effect of parsing the relative clause when the underlying language model is allowed to contextualize against the prior sentence, expecting this to aid the parser in assigning the correct reading. In this setting, we would ideally hope for high accuracy across the board. Finally, we examine the effect of additionally finetuning the parser components on a small amount of training data to see if the parser can pick up on the task.

\subsection{Experimental setup}

We organize the test cases into a train/dev/test split in order to compare the parser baseline against a finetuning setting, where the parser is additionally trained to recognize examples of disambiguating context to learn how to choose the correct interpretation of the relative clause.

\paragraph{Data preparation} In order to not allow overfitting of the model -- we feed it data that is engineered to be task-specific -- we select a small amount of training data against larger development and test sets. We separate the verbs involved in the three different data sets as another measure to avoid overfitting. This leads to a training set of 2640 samples, against a development set of 4400 samples and a test set of 5556 samples.

\paragraph{Parser setup} We initially train both parsers on the original examples from Lassy-Small, as the neural proof net parser was trained on this. However, plain evaluation on the test data is not an option given that we want to prepend a prior disambiguating sentence; the fact that BERT employs positional embeddings makes the parsers unsuitable for the task. Hence we train the parsers from scratch, prepending a random number of unattended tokens, ranging between 5-80 tokens. This ensures that the parser will be robust against the position shifting in later experiments. Due to the difference in parsing regime, baseline scores for the position-shifted parsers follow different evaluation metrics, displayed in Table \ref{table:parser_accuracy}.

\begin{table}[h!]
    \centering
    \begin{tabular}{@{}ccccc@{}}
        \toprule
            \multicolumn{2}{c}{\emph{NPN}} & & \multicolumn{2}{c}{\emph{UD}} \\
            \midrule
         92.98 & 54.87 & & 88.37 & 86.92 \\
        \bottomrule
    \end{tabular}
    \caption{Baseline parser accuracy scores.}
    \label{table:parser_accuracy}
\end{table}

For the neural proof net parser we report tagging accuracy (percentage of total supertags correctly predicted) and frame accuracy (percentage of sentences for which all supertags were correctly predicted). These numbers are only slightly lower than the original parser.\footnote{Tagging accuracy: 93.21, frame accuracy: 56.36} For the UD parser, we compute unlabelled and labelled attachment score, which are comparable to state of the art.\footnote{The Spacy Dutch UD parser \texttt{nl\_core\_news\_lg} reports a UAS score of 87 and a LAS score of 83 (\url{https://spacy.io/models/nl})}

\subsection{First experiment}

For the initial evaluation of the parsers, we assess their disambiguation performance in two scenarios: first, we ask the parsers to parse only the ambiguous phrase, inspecting the initial bias the parsers have obtained from their training data. In the next experiment, we allow the underlying language model to contextualize against the disambiguating prior sentence, and we let the parser component then parse the ambiguous phrase to see if it can succesfully exploit the LM's contextualization capabilities.

Table \ref{table:results_main} displays the results for the first scenario, where we present the relative clause in two possible orders: one in which the regular order is presented (\textbf{S Pron O V}), and one in which subject and object are interchanged (\textbf{O Pron S V}). In the case of irreversible triples, this means that in the regular order we expect a high-performing parser to always assign the subject-relative interpretation, but for the reversed order we expect the object-relative interpretation. For the reversible cases, we expect a 50/50 accuracy for both presentations if there is a weak preference for either order, and a skew towards the subject-relative interpretation in the case of strong lexical preference.

\begin{table}[h!]
    \centering
    \begin{tabular}{@{}l@{\hskip 0.75em}c@{\hskip 0.75em}c@{}}
    \toprule
       \emph{Neural Proof Nets}            & \textbf{S die O V} & \textbf{O die S V}  \\
                                    & (\emph{subj-rel}) & (\emph{obj-rel})  \\
     \midrule
     \textbf{Irreversible}      & 98.76   & 61.86 \\
     \textbf{Reversible-strong} & 95.37 &  2.32 \\
     \textbf{Reversible-weak}   & 97.64 &  1.63  \\
    \toprule
       \emph{Universal Dependencies}            & \textbf{S die O V} & \textbf{O die S V}  \\
                                    & (\emph{subj-rel}) & (\emph{obj-rel})  \\
     \midrule
     \textbf{Irreversible}                  & 95.72  & 27.85 \\
     \textbf{Reversible, Strong pref.}      & 94.88  & 0.65 \\
     \textbf{Reversible, Weak pref.}        & 92.03  & 0.26 \\
    \bottomrule
    \end{tabular}
    \caption{Accuracy results for three different relative clauses without context. Left: presenting the relative clause in regular word order. Right: presenting the relative clause in reversed order. These results indicate the baseline parsing preference without any disambiguating prior sentence.}
    \label{table:results_main}
\end{table}

The result displays a clear preference for the subject-relative interpretation. In the case of irreversible triples the parsers both pick up on the fact that presented a reversed order must obtain the object-relative interpretation. On the other hand, the results for the reversible triples are significantly below expectation, in the sense that regardless of the presented word order, they will almost always assign a subject-relative interpretation ($>95\%$ on the left, $<6\%$ on the right).

These results are somewhat to be expected: the subject-relative this reading prevails in the training data that the parsers were trained on: a total of 306 cases of the subject-relative interpretation occur in Lassy Small, versus 32 cases of the object-relative interpretation. One could then argue that this is indeed the natural intended interpretation so it should have been picked up by any parser replicating its training data, explaining the results in Table \ref{table:results_main}.

\paragraph{Contextualization} However, if it were the case that the parser can easily exploit the information embedded in the language model, we would expect to see that setting the model such that the prior sentence is indeed attended to by the underlying BERT model, the performance would increase. Table \ref{table:results_context} displays the results for this second scenario. Here, by introducing the prior sentence as disambiguating context, the ideal parser scores upward to $100\%$ everywhere, thus indicating it can make the correct parse in context.

\begin{table}[h!]
    \centering
    \begin{tabular}{@{}l@{\hskip 0.75em}c@{\hskip 0.75em}c@{\hskip 0.75em}c@{\hskip 0.75em}c@{}}
    \toprule
      First                 & {\small \textbf{SVO}} & {\small \textbf{SVO}} & {\small \textbf{OVS}} & {\small \textbf{OVS}} \\
      Second                & {\small \textbf{S die O V}} & {\small \textbf{O die S V}} & {\small \textbf{S die O V}} & {\small \textbf{O die S V}} \\
      Reading         & {\small (\emph{subj-rel})} & {\small (\emph{obj-rel})}  & {\small (\emph{obj-rel})} & {\small (\emph{subj-rel})} \\
     \midrule
        \emph{NPN} & & & \\
        \textbf{Irrev.}  & 98.88 & 69.77  &  N/A & N/A \\
        \textbf{Strong} & 98.98 & 10.13 & 1.81 & 93.05 \\
        \textbf{Weak}   & 98.83 &  4.76 & 2.14 & 97.19 \\
    \toprule
        \emph{UD} & & & \\
        \textbf{Irrev.} & 96.91   & 35.03   & N/A & N/A \\
        \textbf{Strong} & 96.61   &  1.29  & 0.51 & 96.06   \\
        \textbf{Weak}   & 94.77   &  0.46 & 0.45 & 94.30 \\
    \bottomrule
    \end{tabular}
   \caption{Accuracy results for three different relative clauses with context, i.e. the prior sentence is attended to by the language model prior to parsing. These results indicate the effect of contextualizing on parsing disambiguation performance.}
    \label{table:results_context}
\end{table}

We observe a stable accuracy for the cases of subject-relative readings, with significant accuracy gains for the object-relative reading that is less persistent in the original training data. This shows that, to some extent, the added information of the underlying language model gives the parser the incentive to pick up on the grammatical relations in the prior sentence. However, the results are not particularly encouraging: while both parsers do increase in accuracy overall, their strong bias towards a subject-relative interpretation remains.

\subsection{Finetuning} After evaluation of the influence of the prior sentence through the contextualization of the BERT embeddings, we additionally finetune the parsers on our task, to see whether the parser could in principle assign the correct interpretation given that the underlying BERT embeddings have access to the prior sentence for contextualization. We explicitly do not update the language model itself, as we want to investigate the parser's capacity for disambiguation, and allowing the language model to update would be too prone to overfitting \cite{rogers-etal-2020-primer}.

The finetuning scenario thus serves as a means to measure the extent to which the parsers' strong structural bias can be mitigated. Given the fact that the relative clause has an unambiguous interpretation in the contextualized scenario, training is straightforward, and the results reflect whether the parser by itself can pick up on the contextualized lexical information provided by the language model. These results are displayed in Table \ref{table:results_context_finetune}.

\begin{table}[h!]
    \centering
    \begin{tabular}{@{}l@{\hskip 0.75em}c@{\hskip 0.75em}c@{\hskip 0.75em}c@{\hskip 0.75em}c@{}}
    \toprule
      First                 & {\small \textbf{SVO}} & {\small \textbf{SVO}} & {\small \textbf{OVS}} & {\small \textbf{OVS}} \\
      Second                & {\small \textbf{S die O V}} & {\small \textbf{O die S V}} & {\small \textbf{S die O V}} & {\small \textbf{O die S V}} \\
      Reading         & {\small (\emph{subj-rel})} & {\small (\emph{obj-rel})}  & {\small (\emph{obj-rel})} & {\small (\emph{subj-rel})} \\
     \midrule
        \emph{NPN} & & & \\
        \textbf{Irrev.} & 89.98 & 91.67   & N/A & N/A \\
        \textbf{Strong} & 65.64 & 74.45 & 40.02 & 30.98  \\
        \textbf{Weak}   & 61.31 & 64.43   & 46.59 & 45.55 \\
    \toprule
        \emph{UD} & & & \\
             \textbf{Irrev.} & 89.95 & 71.89 & N/A & N/A  \\
             \textbf{Strong} & 72.92 & 24.43 & 19.71 & 65.56 \\
            \textbf{Weak}   & 67.95  & 17.63 & 14.43 & 64.52 \\
    \bottomrule
    \end{tabular}
    \caption{Accuracy results for three different relative clauses with context, i.e. the prior sentence is attended to by the language model prior to parsing, after finetuning different parts of the parser model on the task itself.}
    \label{table:results_context_finetune}
\end{table}

In these results we observe that it is indeed possible to leverage the training task to even out the parsers' bias through grounding, leaving the neural proof net parser accuracy evenly distributed over cases of subject-relative and object-relative interpretations. On the other hand the UD parser does not adapt to the task that well and retains the strong bias toward a subject-relative interpretation. Overall we observe that the price one pays for the increased accuracy in cases of the object-relative interpretation, is a signficant drop of performance in the subject-relative case, showing that developing a balanced parser is no easy task.

\section{Discussion}

In the setup of our experiments, we were careful to develop test cases that target the parsers in a few different settings. Rather than expecting high performance overall, the aim of the experiment is to both measure the prevalence of structural bias in the parser, as well as measuring to what extent such bias can be mitigated, if present. As such, the experimental results shouldn't be taken as proof that the parsers in themselves are necessarily insufficient.

Rather, it should be taken as a point to argue that parsers should generally take ambiguity into account, and while lexical ambiguity can be addressed by means of a neural model iterating over a balanced training dataset, the inability to accommodate syntactic ambiguity, both in the training corpora used, as well as in the parser architectures involved, poses a problem that our experiments confirm.

Aside from the methodological viewpoint above, one may argue that we need not care about the presented experiment as we could simply finetune underlying language model together with the parser on top and achieve high performance on the experiments. While it is true that finetuning the BERT model alongside the parser leads to peak performance, this is most likely due to the language model picking up on positional information quickly, and not due to the model becoming a better parser.
To back this claim, we measure parser accuracy metrics on the original training corpus, for finetuned models that only adapt the parser or include the underlying language model. The metrics are displayed in Table \ref{table:parser_accuracy_after_finetuning}.

\begin{table}[h!]
    \centering
    \begin{tabular}{@{}lccccc@{}}
        \toprule
            & \multicolumn{2}{c}{\emph{NPN}} & & \multicolumn{2}{c}{\emph{UD}} \\
            \midrule
         & Tag & Frame & & UAS & LAS  \\
         \textbf{Parser only} & 72.54 & 10.32 & & 87.41 & 85.92 \\
         \textbf{LM+Parser}   & 73.13 & 9.77 & & 80.39 & 78.58 \\
        \bottomrule
    \end{tabular}
    \caption{Parser accuracies on the test set of Lassy Small, after finetuning on the disambiguation experiment. The top row gives the results for the actual finetuned models that we evaluate, where the bottom row indicates accuracies for models where the language model is included in the finetuning process.}
    \label{table:parser_accuracy_after_finetuning}
\end{table}

For the neural proof net parser, we observe that performance drastically declines in both cases, with larger decline for the case where the language model's parameter were included in the finetuning process. The UD parser on the other hand does not decrease in performance so much, but has a strong decline once the language model is included.
This highlights two points: first, given that the neural proof net parser adapted itself better to the task of our main experiment, we conclude that the price to pay for task adaptation is a decline in overall parser performance. Second, we argue that including the language model in finetuning leads to overfitting on the task, and reduces the parser's overall accuracy.


\section{Conclusion}

By introducing a synthetic test set of naturalistic Dutch relative clauses, we carried out an experiment to investigate the sensitivity to structural bias of two parsing architectures that are both based on a BERT-style language model. The experiments show that both parser pick up on structural preference for a subject-relative reading of the relative clause, following a strong statistical bias coming from the data they were trained on. Further experimentation shows that a more complex neurosymbolic parsing regime adapts more easily to a bias correcting finetuning setup than a universal dependencies parser implemented as a sequence labelling model. However, in both cases performance on the task is severely below expectation, and we hope that this work inspires further work on careful data augmentation and parser development.



\bibliographystyle{acl_natbib}
\bibliography{anthology,custom}


\end{document}